\documentclass[conference]{IEEEtran}
\IEEEoverridecommandlockouts
\usepackage{amsmath,amssymb,amsfonts}
\usepackage{algorithmic}
\usepackage{graphicx}
\usepackage{textcomp,hyperref}
\usepackage{xcolor}
\usepackage{blindtext}
\usepackage[
    backend=biber,
    style=chem-angew,
  ]{biblatex}
  \newcommand{\linebreakand}{%
  \end{@IEEEauthorhalign}
  \hfill\mbox{}\par
  \mbox{}\hfill\begin{@IEEEauthorhalign}
}

\title{Real Time Human Detection by Unmanned Aerial Vehicles}

\author{
\IEEEauthorblockN{Walid Guettala}
\IEEEauthorblockA{Computer Science Department, \\
Biskra University, Algeria \\
walidguettala@gmail.com}
\and
\IEEEauthorblockN{Ali Sayah}
\IEEEauthorblockA{Computer Science Department, \\
Biskra University, Algeria \\
Sayah.Ali@hotmail.com}
\and
\IEEEauthorblockN{Laid Kahloul}
\IEEEauthorblockA{LINFI Laboratory, Computer Science Department, \\
Biskra University, Algeria \\
l.kahloul@univ-biskra.dz}
\and
\IEEEauthorblockN{Ahmed Tibermacine}
\IEEEauthorblockA{LESIA laboratory, Computer Science Department, \\
Biskra University, Algeria \\
ahmed.tibermacine@univ-biskra.dz}
}

\begin{document}

\maketitle

\begin{abstract}
One of the most important problems in computer vision and remote sensing is object detection, which identifies particular categories of diverse things in pictures. Two crucial data sources for public security are the thermal infrared (TIR) remote sensing multi-scenario photos and videos produced by unmanned aerial vehicles (UAVs). Due to the small scale of the target, complex scene information, low resolution relative to the viewable videos, and dearth of publicly available labeled datasets and training models, their object detection procedure is still difficult. A UAV TIR object detection framework for pictures and videos is suggested in this study. The Forward-looking Infrared (FLIR) cameras used to gather ground-based TIR photos and videos are used to create the ``You Only Look Once'' (YOLO) model, which is based on CNN architecture. Results indicated that in the validating task, detecting human object had an average precision at IOU (Intersection over Union) = 0.5, which was 72.5\%, using YOLOv7 (YOLO version 7) state of the art model \cite{1}, while the detection speed around 161 frames per second (FPS/second). The usefulness of the YOLO architecture is demonstrated in the application, which evaluates the cross-detection performance of people in UAV TIR videos under a YOLOv7 model in terms of the various UAVs' observation angles. The qualitative and quantitative evaluation of object detection from TIR pictures and videos using deep-learning models is supported favorably by this work.
\end{abstract}


\section{Introduction}
Unmanned aerial vehicle (UAV) object detection is developing technology with a wide range of uses, including aerial picture analysis, intelligent surveillance, and route inspection \cite{2,3}. Recently, there has been a lot of advancement in object detection. The deep neural network (DNN), in particular the convolutional neural network (CNN) \cite{4}, has shown record breaking performance in computer vision applications like object recognition \cite{5}, especially with the introduction of large-scale visual datasets and greater computing power. However, given the unique perspective, it is still a difficult task.

``Deep-learning-based object detection'' \cite{6} and ``conventional manual feature-based object detection'' \cite{7} are two different approaches to object detection. It focuses on the target-feature extraction technique design for manual feature-based object detection, but because it is still difficult to meet various constraints, most of these sorts of approaches are only employed in certain environments \cite{8}. On the other hand, deep-learning-based techniques may now achieve real-time detection in addition to improving accuracy as computing technology advances.

Although deep-learning-based techniques have significantly advanced object recognition, miss-detection problems still exist in UAV. The following are the key contributing factors to these problems: (1) The network's receptive field is not sufficiently resistant to small objects and Thermal Imaging, and (2) the training dataset is confined to UAV viewpoint. In general, object feature representation and the associated training dataset are crucial for enhancing object detection performance. Additionally, the trade-off between accuracy and processing speed is crucial for real-world applications.

We are motivated by these issues to create an object detection technique based on the You Only Look principle (YOLO) \cite{9}, and we focus on the detection of tiny objects. To enhance detection performance on tiny objects, we gather data based on UAV views, and we enhance the YOLOv7 network. to our dataset by transfer learning. The following are some of our study's contributions: (1) develop a UAV perspective-based dataset for person detection that may be used to enhance human detection; (2) enhance YOLO's network architecture to expand the receptive area and further improve tiny human detecting performance using transfer learning.

The remaining of this article is structured as follows: The related work is introduced in Section 2, the experimental setup is further explained in Section 3, and presents the experimental findings and talks about detailed comparative analysis. Concluding observations are included in Section 4.

\section{Related work}
In the literature, considerable numbers of works has been introduced to handle the challenging tasks of object detection. This section will briefly discuss novel approaches and methods.

This work is associated with developing a human detection system based on YOLO v4, utilizing the single shot detection idea \cite{10}. This approach does not follow semantic segmentation, it does not experience loss of the object information, such as the disappearance of the gradients, and it does not call for pre-defined anchors, in contrast to second stage region-based object detection systems. This method is particularly rapid in multi-threaded GPU systems and includes robust feature extractors and reinforced multi scale object detection. Since the focus of their foundational study is object classification for Unmanned Aerial Vehicle (UAV) applications, they initially decide to identify people using thermal datasets. As a result,  to develop and test the model, they used thermal photos and videos obtained from UAV thermal cameras that were flying 1 to 50 meters above the ground. To extract features such as Weighted Residual Connections (WRC), Cross Stage Partial Connections (CSP), Cross mini Batch Normalization (CmBN), Self-Adversarial Training (SAT), Mish Activation (MA), Mosaic Data Augmentation (MDA), and Drop Block Regularization, the YOLO v4 uses ground truth bounding boxes (DBR). The modeling employing YOLO v4 performs in a promising manner, according to the best performance of these models in terms of mean Average Precision (mAP) around 48\%.
 
In this paper, the tracking system is composed of two stages to address the challenges of tracking pedestrians \cite{11}. Pedestrians are identified by a pedestrian detection method in the first stage, and their trajectories are estimated in the second stage to assist the navigation of the UAV. The pedestrian detection method is composed of two processes: feature extraction and classification. In the feature extraction process, HOG and DCT are employed to extract pedestrian shape and texture features, respectively. These features are subsequently analyzed by a Support Vector Machine (SVM) to classify pedestrians. A fusion of the HOG and DCT descriptors is used to improve the accuracy of pedestrian detection.

The surveillance system described in this work is based on infrared video and has resolution-enhanced automated target detection/recognition (ATD/R) technology \cite{12} that is extensively used in both military and civilian applications. A super-resolution approach is used to raise target signature resolution and optimize the baseline quality of inputs for object detection to address the problem of tiny numbers of pixels on target in the developed ATD/R system, which are encountered in long range imaging. We train a sophisticated and potent convolutional neural network (CNN) based faster-RCNN utilizing in-house generated and labeled long wave infrared images datasets to take on the issue of recognizing extremely low-resolution objects. Two datasets with target categories including pedestrians and six different types of ground vehicles were used to test the system under various weather situations. The designed ATD/R system successfully addresses the low small number of pixels on target, faced in long range applications, and can detect extremely low-resolution targets with improved performance.

On the one hand, we generate a dataset utilized for network training and testing with the intention of establishing a quick UAV viewpoint approach for specific purposes of object detection. On the other side, we provide a YOLOv7-based UAV-viewed object detection approach. To enhance YOLOv7's effectiveness while detecting small objects, most articles employ small distance human detection rather than far distance person detection, as shown in figure 2. Then, using data augmentation, the model is trained using the obtained data. while training. On the basis of maintaining the original YOLOv7 performance, we test our model using our collected test dataset. Our suggested model may further improve the performance of tiny-human object detection. 

\section{Experimental setup} 

In our system project, two entities are specified in the proposed system to carry out the thermal video stream-based human detection. These are as follows: 1) the controller of this UAV by human, and 3) a small platform (Google Colab) executes the proposed algorithm to process the video and detect humans.

\subsection{Proposed Method} 
YOLOv7 trainable bag of freebies sets, the paper states that the model can efficiently predict video inputs ranging from 5 fps to 160 fps YOLOv7 has the highest average precision of 56.8\%, the comparison with YOLOv4 \cite{13}, the cost of running the model has been reduced by 50\%, the parameters in the hidden layer are also reduced up to 40\%, YOLOv7 has achieved 1.5 times higher average precision than YOLOv4, YOLO uses a convolution neural network layers \cite{14} to predict bounding boxes and class probabilities considering the entire image in a single evaluation in one step and for one unit, YOLO predicts multiple bounding boxes the class probabilities for each box and all the bounding boxes across the classes making it the one stage detection model unlike earlier object detection models which localize objects and images by using regions of the image with high probabilities of contender YOLO considers the full image.

\begin{figure*}[h]
\includegraphics[width=0.9\linewidth, height=6cm]{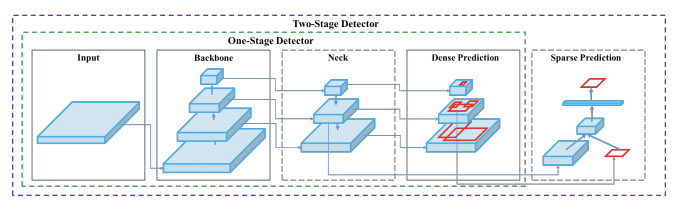}

\caption{YOLO architecture}
\end{figure*}

\subsubsection{Architectural reforms}  
 Now we'll talk about the architecture of YOLO image frames are featured through a backbone which is then combined and mixed in the neck and then they are pass along and YOLO predicts the bounding boxes the classes of the bounding box and objects of the bounding boxes as figure 1 shows, let's understand each of its modules separately first we have the input layer is nothing but the image, then  the backbone it's a deep neural network composed mainly of convolutional layers ,the main objective of the backbone is to extract the essential features the selection of backbone is a key step as it will improve the performance of object detection, often pre-trained neural networks are used to train the backbone some of the commonly used pre-trained networks are VGG-16 \cite{15}, RetinaNet \cite{16} resnet50 \cite{17} etc, for YOLOv7 the paper used the following pre-trained weights VoVNet \cite{18} and ELAN, next object detector models insert additional layers between backbone and head which are referred to as a copy of the detectors the essential role of neck is to collect feature maps from different stages, usually a neck is composed of several bottom-up parts and several top-down parts for enhancement  detection happens, in the head so the head is also called as dense prediction to set the director to decouple the object localization and classification task for each module once the detectors make the prediction for this localization and classification at the same time, this layer is present only in one stage after detectors like YOLO, Single Shot Detector(SSD) \cite{19} into self-detection they were completely different sparse prediction is for two-stage detectors Faster R-CNN \cite{20} and R-FCN \cite{21} highly different, which does the traditional class probabilities for the model input, our YOLO is one stage together they form the YOLO architecture 

Let's dive deeper into the topics and technical words that we talked before the first term is bag of freebies bag of freebies model, refers to increasing the model accuracy by making improvements without actually increasing the training cost the older versions of YOLO, bag of freebies used for this particular paper (YOLOv7), batch normalization \cite{22} the purpose of this layer is to integrate the mean and variance of batch normalization into the bias and rate of the convolution layer, second implicit knowledge in combined with convolutional feature maps, the final EMA model, EMA is the technique used in mean teacher and we use the EMA model purely as the final interference model training optimizer. The author uses gradient prediction to generate course-defined hierarchical labels, the author also used extended efficient layer aggregation networks, the author also performed a model scaling for concatenation based models and identifying connections in one convolutional layer, and finally, the author used COCO dataset to train the YOLOv7 from scratch. The efficiency of a YOLO network's convolutional layers in the backbone is essential to efficient interference speed, the authors built and researched that the distance it takes the gradient to back propagate through the layers the shorter the gradient the more powerful the network will be to learn, the final layer aggregation they choose is E-ELAN in an extended version of the ELAN increasing the accuracy as the model is now capable of identifying small objects and large objects the scaling factors, re-parameterizing uses gradient flow propagation plans to analyze how re-parameterized convolutions should be combined with different network re-parameterization technique involves averaging a set of model weights to create a model that is more robust to the general pattern that is trying to model.

\subsection{Dataset}
The UAV was used to fly over the Scottish wilderness in South Africa to record videos and gather data. 
The business Aerialworks assisted in collecting data from various altitudes and up to 75 meters away from the human object. 
The 640 x 512 pixel video images were taken out to be processed further. 
3,807 annotated photos totaling 2,358 positive and 1449 negative images made up the dataset, which was compiled from 7 video segments. 
Each image on the Roboflow website was individually annotated by the authors, resized to 640 x 640, and then data augmentation (flip, corp, rotate 90°) was applied to obtain around 9369 photos in total. 

\subsection{Training}
The training was performed on the free version of Google Colab, it allows anybody to write and execute arbitrary python code through the browser, and is especially well suited to machine learning, data analysis and education. More technically, the system is equipped with: 
NVIDIA Tesla K80, 2-core Xeon 2.2GHz, 13GB RAM, 33GB HDD and Python 3.6. We determined the following hyperparameter values to train our detection system:
The batch size is set to 16, the initial learning rate is set at 0.01 with weight decay and momentum set to 0.937 and 0.0005, respectively, and the number of iterations is 40.
640 × 640 is the new size for testing photos.

\subsection{Expermintal Analyse  and results}
This study gathers much more images with UAV and shooting by Drone Guards company because there aren't many UAV perspective datasets online. 
It contains 9369 images in total, divided into training data (8300 images), valid (679 images) and test data (347 images). 
Following three guidelines, they are classified into training, valid, and test data: 
(1) Images from the same video are gathered discretely, which means that we take one picture per second. (2) Images from the same video can either be training data, valid data, or test data. (3) Each video is taken from a distinct trip place. 

We utilize transfer learning to train the model YOLOv7 from the Github link \href{https://github.com/WongKinYiu/YOLOv7}{YOLOv7} as we are mostly interested in human detection.

\begin{figure}[!htb]
    \centering
    \includegraphics[width=0.44\textwidth,height=6cm]{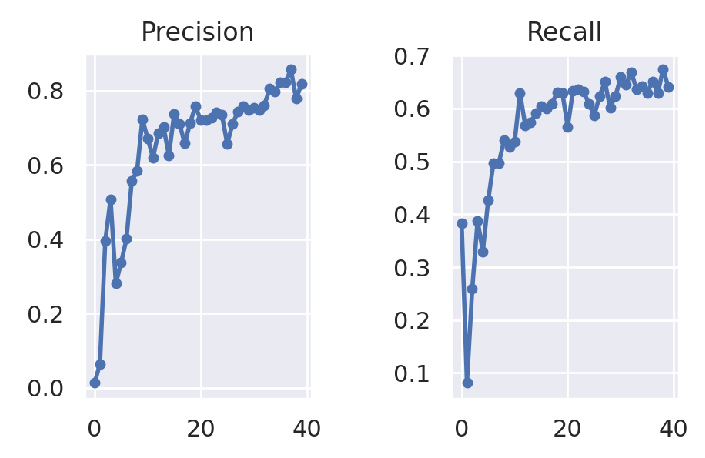}
    \hspace{0.02\textwidth} 
    \includegraphics[width=0.4\textwidth,height=6cm]{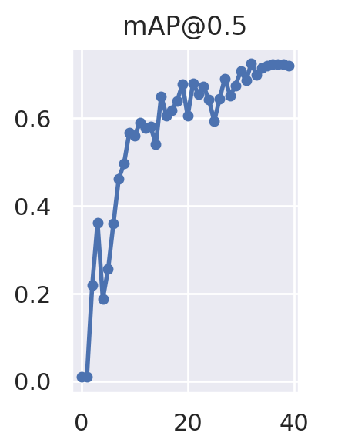}
    \caption{Precision, Recall, and mAP@0.5 on validation dataset during training}
\end{figure}

\begin{table*}[!htb]
\caption{mAP performance comparison on different detectors}
\begin{tabular}{|c|c|c|c|c|}
\hline
Ref & Algorithm & Exec. Environment & mAP\% & Speed (FPS/second) \\
\hline
1 & YOLOv3 & PC & 61.72 & 20 \\
2 & UAV-YOLO & PC & 64.42 & 20 \\
3 & SSD512 & PC & 56.01 & 23 \\
4 & model2 & Jetson AGX Xavier & 89.26 & 24.6 \\
5 & Our-YOLOV7 & Google-Colab & 72.50 & 161 \\
\hline
\end{tabular}
\label{table:nonlin}
\end{table*}

\begin{table}[!htb]
\caption{mAP performance comparison on different detectors}
\centering
\begin{tabular}{|c|c|c|c|c|}
\hline
Ref & Algorithm & Exec. Environment & mAP\% & Speed (FPS/second) \\
\hline
1 & YOLOv3 & PC & 61.72 & 20 \\
2 & UAV-YOLO & PC & 64.42 & 20 \\
3 & SSD512 & PC & 56.01 & 23 \\
4 & model2 & Jetson AGX Xavier & 89.26 & 24.6 \\
5 & Our-YOLOV7 & Google-Colab & 72.50 & 161 \\
\hline
\end{tabular}
\label{table:nonlin}
\end{table}

the values of mAP, precision, and recall are shown in figure 2. We can see that mAP remains stable throughout the last 10 iterations, suggesting that the model converges while keeping precision and recall around 81.7\% and 64\%, respectively, and our best iteration mAP scores 72.5\% with precision of 76\% and recall score around 66.8\%.

\begin{figure}
\includegraphics[width=0.9\linewidth, height=6cm]{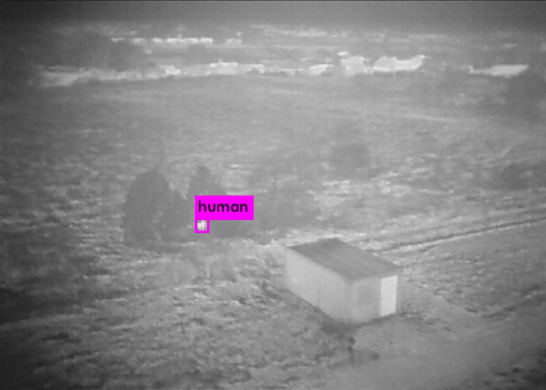} 

Sample Image from  model2 Dataset.

\includegraphics[width=0.9\linewidth, height=6cm]{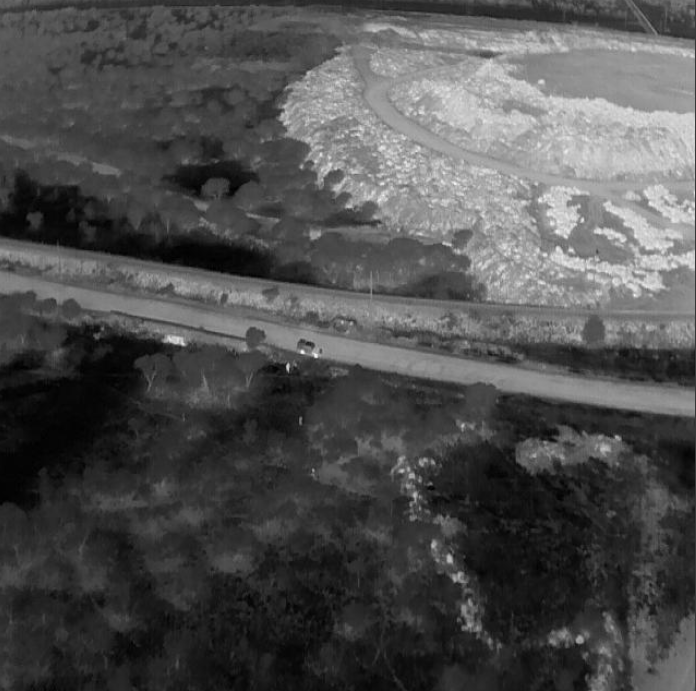}

Sample Image from  our Dataset.
\caption{Comparison between our dataset sample and Model2 dataset sample}
\end{figure}

\begin{figure*}
    
    \includegraphics[width=.33\textwidth,height=6cm]{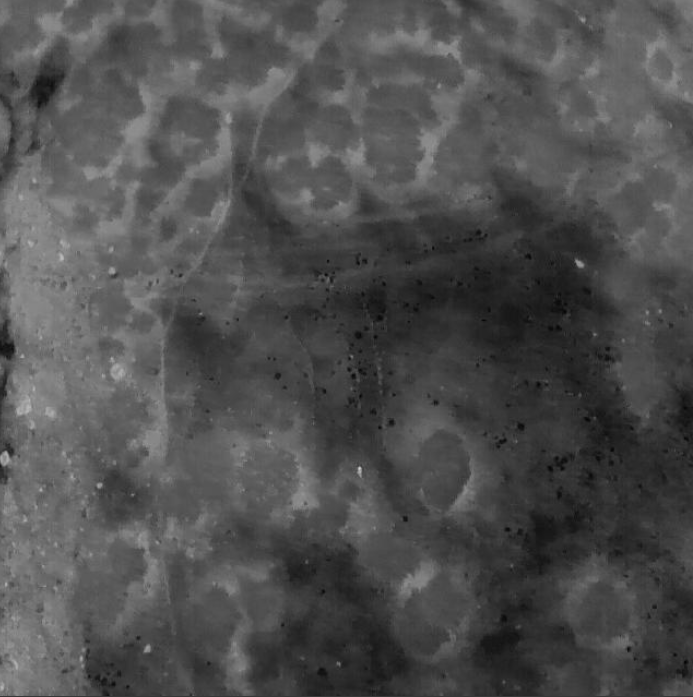}
    \includegraphics[width=.33\textwidth,height=6cm]{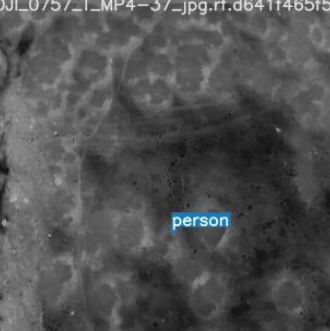}
    \includegraphics[width=.33\textwidth,height=6cm]{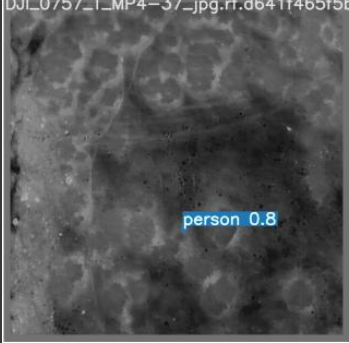}
    \\[\smallskipamount]
    \includegraphics[width=.33\textwidth,height=6cm]{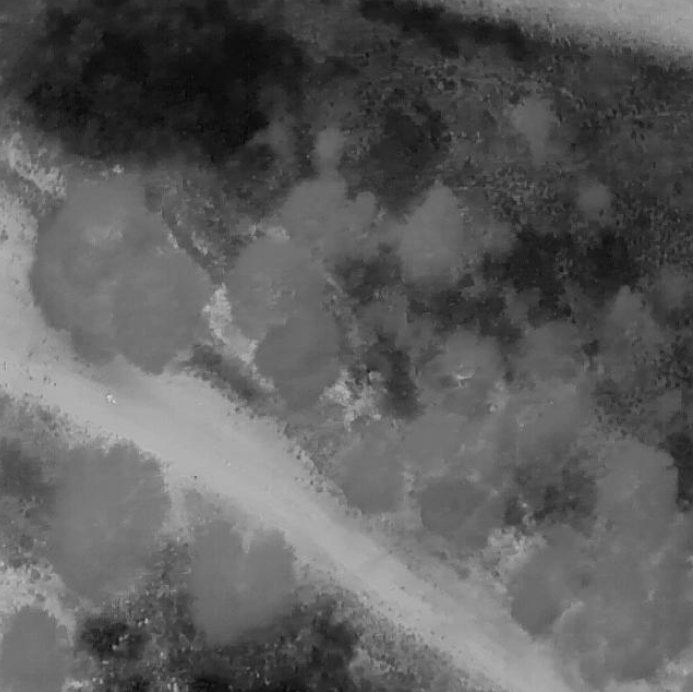}
    \includegraphics[width=.33\textwidth,height=6cm]{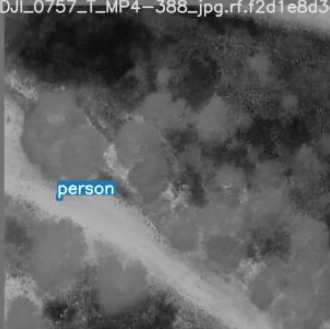}
    \includegraphics[width=.33\textwidth,height=6cm]{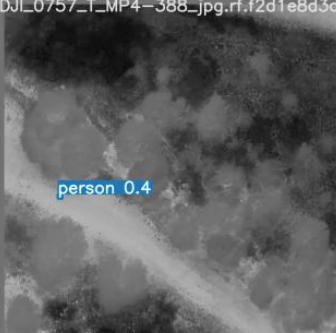}
    \\[\smallskipamount]
    \includegraphics[width=.33\textwidth,height=6cm]{4.png}
    \includegraphics[width=.33\textwidth,height=6cm]{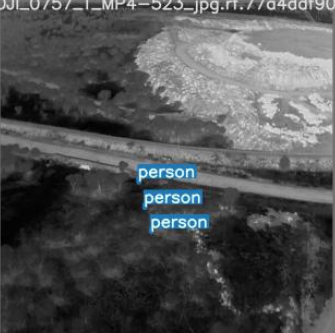}
    \includegraphics[width=.33\textwidth,height=6cm]{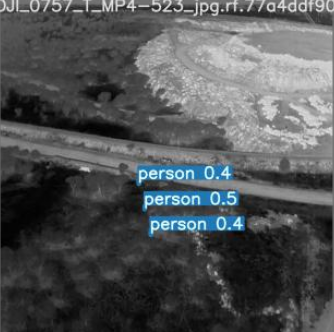}
    
    Sample
    \hfill
    Labeled
    \hfill
    Predicted
    
    \caption{Sample images from the stream, labeled and predicted images}\label{fig:foobar}
\end{figure*}

It is evident that  model2 \cite{23} beats our YOLOv7 by getting better to mean average precision (mAP) however our model outperforms the UAV-YOLO model \cite{24}, and this outcome was influenced by the type of data used and tested by each model because they were trained on different datasets (UAV-YOLO: RGB image and small distance from the object, model2: thermal image with clear human from not far distance), as you can see Figure 3 shows in the picture of model2 example image from their dataset compared to our image. Although you can see that they utilized superior computers to train their models, we have double the speed (fps/second) of their model despite the fact that the difference in object clarity in the image, distance from the camera, and complicated background we have in our dataset (Figure 4 image samples from our dataset) might have a significant impact on our model's performance despite these facts our approach get good results like image represented in Figure 4, first row image and 3rd image have a complex background with tiny person hard even for a person to recognize them, second row image small objects with the same background color of human hard to figure out where is the person,  our manual labeling we used video to help us see the moving object (human) to label them in a good manner.
An illustration of the detected results of our YOLOv7 model for some samples is shown in Figure 2.From a vision perspective, the YOLOv7 model can tackle miss-detection and false detection

\section{Conclusion and future works}
In this paper, we propose a UAV perspective object detection method based on YOLOv7.We analyze and find the reason causing small object miss-detection and false detection is limited receptive field We gather data that is specifically utilized for small object detection from a UAV perspective. Based on the UAV dataset, we trained our model to recognize human objects in thermal vision using data collected by a UAV.  Representative experimental results comparing our method to competing approaches in a range of difficulties show that our method generally performs well against competing approaches in a variety of difficulties, particularly in small object detection. It shows that The YOLOv7 performs admirably when trying to find small items in difficult environments.

In the future, we will use various versions of YOLOv7 and make use of our videos recorded by UAV camera to create two more classes for the detection of cars and animals, as well as more data to test the human object detection model. On the other hand, we will try to implement object detection on videos by outlining the path that an object took to help us detect and observe how the object moved using DeepSORT or StrongSORT on top of our object detection model.

\section{Acknowledgments}
We thank all the people that have made work  come to realization. 

\printbibliography
\vspace{12pt}

\end{document}